\pdfoutput=1

\documentclass[11pt]{article}
\usepackage{coling2020}
\usepackage{times}
\usepackage{url}
\usepackage{latexsym}

\usepackage{microtype}
\usepackage{graphicx}
\usepackage{wrapfig}
\usepackage{amssymb}
\usepackage{color,xcolor,colortbl}
\usepackage{enumitem}
\usepackage{soul}
\usepackage{amsmath}
\usepackage{url}
\usepackage{algorithm}
\usepackage{algorithmic}
\usepackage[font={small}]{caption}
\usepackage{booktabs}
\usepackage{bm,bbm}
\usepackage{mathtools}
\usepackage{array}
\usepackage{multirow}
\usepackage{subcaption}
\usepackage{pbox}
\usepackage{pifont}
\usepackage{arydshln}
\usepackage{dblfloatfix}
\usepackage{relsize}

\definecolor{darkblue}{rgb}{0.0, 0.0, 0.55}
\newenvironment{fontppl}{\fontfamily{ppl}\selectfont}{\par} 
\setulcolor{blue} 
\usepackage{soul}

\definecolor{d}{HTML}{c6dbef}
\definecolor{l}{HTML}{e5f5e0}

\usepackage{pifont}
\usepackage[T1]{fontenc}
\usepackage[scaled=.8]{beramono}

\colingfinalcopy 

\title{How Domain Terminology Affects Meeting Summarization Performance}

\author{Jia Jin Koay, Alexander Roustai, Xiaojin Dai, Dillon Burns, Alec Kerrigan, Fei Liu\\[0.5em]
Computer Science Department, University of Central Florida\\
Orlando, FL 32816, USA\\[0.5em]
\texttt{\{jjkoay,alexroustai,xd.zangyiwu,dburns9898,aleckerrigan\}@knights.ucf.edu}\\
\texttt{feiliu@cs.ucf.edu}
}

\date{}

\begin{document}
\maketitle
\begin{abstract}

Meetings are essential to modern organizations. 
Numerous meetings are held and recorded daily, more than can ever be comprehended.
A meeting summarization system that identifies salient utterances from the transcripts to automatically generate meeting minutes can help. 
It empowers users to rapidly search and sift through large meeting collections. 
To date, the impact of domain terminology on the performance of meeting summarization remains understudied, despite that meetings are rich with domain knowledge. 
In this paper, we create gold-standard annotations for domain terminology on a sizable meeting corpus; they are known as jargon terms.
We then analyze the performance of a meeting summarization system with and without jargon terms.
Our findings reveal that domain terminology can have a substantial impact on summarization performance.
We publicly release all domain terminology to advance research in meeting summarization.\footnote{\url{https://github.com/ucfnlp/meeting-domain-terminology}}

\end{abstract}

\section{Introduction}
\label{intro}


\blfootnote{
    %
    %
    %
    %
    %
    
    \hspace{-0.5cm}  
    This work is licensed under a Creative Commons 
    Attribution 4.0 International License.
    License details:
    \url{http://creativecommons.org/licenses/by/4.0/}.
}

A vast number of meetings are being held and recorded everyday, far more than can ever be comprehended. With this explosion of meetings comes a pressing need to develop summarization techniques to assist in browsing meeting archives~\cite{10.1007/11677482_3,ailomaa-etal-2006-archivus}. 
A meeting summarization system takes a meeting recording and its transcript as input and produces a concise text summary as output, which preserves the most important content of the meeting discussion~\cite{murray-carenini-2008-summarizing,liu-liu-2009-extractive,shang-etal-2018-unsupervised,li-etal-2019-keep}. 
The techniques hold great potential to make large archives of meetings substantially more efficient to browse, search and facilitate information sharing.

We envision an automated summarizer that is capable of generating meeting minutes by identifying salient utterances from transcribed meeting recordings.
Neural text summarization has seen significant progress~\cite{see-etal-2017-get,tan-etal-2017-abstractive,chen-bansal-2018-fast,narayan-etal-2018-dont,lebanoff-etal-2018-adapting,west-etal-2019-bottlesum,liu-lapata-2019-hierarchical,laban-etal-2020-summary}, but most prior work focused on written texts.
In contrast, recent years have seen a growing interest in summarizing spoken texts~\cite{tardy-etal-2020-align}.
Particularly, the characteristics of meetings, domain terminology and limited annotated data pose novel challenges to neural summarization models.
We favor extractive over abstractive models as the latter are prone to hallucinate content that is unfaithful to the input~\cite{kryscinski-etal-2019-neural}.

In this paper, we investigate how domain terminology impacts meeting summarization performance, especially in the context of neural extractive summarization. 
Jargon is the specialized terminology associated with a particular domain~\cite{meyers-etal-2014-jargon}.
It is employed in a communicative context and may not be well understood outside that context. 
Because meetings are usually held among professionals, jargon is ubiquitous in meeting discussions.
In Table~\ref{tab:annotation}, we provide an example of jargon terms identified by human experts.
Without a thorough study of technical jargon in the meeting domain, it is unclear how best to optimize a meeting summarizer to incorporate domain knowledge.

We present an assessment of the meeting summarization performance by comparing models trained with and without jargon.
A collection of jargon terms are meticulously compiled by our expert annotators from a meeting corpus containing multi-party conversations on the topic of speech and signal processing~\cite{1198793}.
Such jargon terms are distinct from speech recognition errors; the latter substitutes one word for another similar-sounding word during automatic transcription.
The users can eliminate transcription errors using a modern interactive transcript editor.
However, there remains a pressing need to understand how domain terminology affects the meeting summarization performance.

Our contributions are twofold.
First, we create gold-standard annotations for domain terminology on a large meeting corpus; they are known as jargon terms.
Prior work has not explored such domain-specific thesauri and thus there is limited knowledge of the target domain.
Second, we analyze the performance of a meeting summarization system with and without jargon.
Due to the nature of sound, such a summarizer is highly desirable to aid users in navigating through meeting recordings.
Our findings suggest that domain terminology has a substantial impact on summarization performance, which should not be overlooked.

\begin{table*}
\setlength{\tabcolsep}{5pt}
\renewcommand{\arraystretch}{1.1}
\centering
\begin{fontppl}
\begin{footnotesize}
\begin{tabular}{|lll|}
\hline
\textbf{Start} & \textbf{End} & \textbf{Spoken Utterance}\\
\hline
\hline
247.255 & 252.672 & with Andreas' help um Andreas put together a sort of no frills recognizer which is uh\\
252.672 & 258.837 & gender-dependent but like no adaptation, \textcolor{magenta}{\textbf{no cross-word models, no trigrams -}}\\
& & \textcolor{magenta}{\textbf{a bigram recognizer}}\\
258.837 & 262.221 & and that's trained on \textcolor{magenta}{\textbf{Switchboard}} which is telephone conversations.\\
263.983 & 267.154 & and thanks to Don's help wh- who - Don took\\
267.154 & 270.431 & the first meeting that Jane had transcribed\\
270.431 & 277.520 & and um you know separated - used the individual channels we segmented it in- into\\
& & the segments that Jane had used\\
277.520 & 279.952 & and uh Don sampled that so -\\
281.374 & 289.611 & um and then we ran up to I guess the first twenty minutes, up to \textcolor{magenta}{\textbf{synch time}} of one two\\
& & zero zero so is that - that's twenty minutes or so?\\
289.611 & 296.601 & Um yeah because I guess there's some, and Don can talk to Jane about this, there's some bug \\
& & in \textcolor{magenta}{\textbf{the actual synch time file}} that\\
\hline
\end{tabular}
\end{footnotesize}
\end{fontppl}
\caption{
A snippet of a human transcript that contains spoken utterances and their start/end times. Domain terminology is in bold.
}
\label{tab:annotation} 
\end{table*}

\section{Data and Annotation}

We extend the ICSI meeting corpus~\cite{1198793} for this study, which contains 75 meetings recorded at the International Computer Science Institute, Berkeley.\footnote{\url{http://www1.icsi.berkeley.edu/Speech/mr/mtgrcdr.html}}
The meetings are primarily between speech group members of ICSI. 
An average meeting lasts an hour and has up to 10 participants.
Each participant wore a close-talking microphone and they sat around a meeting table equipped with far-field microphones.
The corpus is one of the larger resources in this area~\cite{Renals:2012}.
It contains rich annotations including human transcripts, segmentation of utterances and further annotations of extractive summaries\footnote{\url{http://groups.inf.ed.ac.uk/ami/icsi/}}, making the corpus suitable for summarization.
We have chosen ICSI over the AMI corpus~\cite{10.1007/11677482_3}; both are natural conversations, but the scenarios in AMI meetings are artificial.

Annotating \emph{domain terminology} is non-trivial as there lacks a universal definition.
Instead, we solicit annotations from undergraduate students majoring in computer science and designate words and expressions that are beyond the scope of their knowledge as domain terminology. 
Interestingly, modern deep neural models often acquire such generic knowledge through unsupervised pretraining~\cite{lewis-etal-2020-bart}.
The annotators are instructed to identify words and expressions from human transcripts;
they are called jargon terms and usually have particular meaning in the speech and language processing field.

The student annotators are able to annotate all of the 75 meetings for jargon terms.
Meeting transcripts are substantially longer than typical news articles.
A transcript contains 1,731 utterances on average and 7 words per utterance.
Each meeting is annotated by one student due to the sheer size of the transcripts. 
However, one of the meetings has been annotated by all of the four annotators.
Their average pairwise inter-annotator agreement is 0.69, indicating a moderate to high agreement between the annotators.
We find that an average meeting contains 92 jargon expressions and each expression contains about 3 words.
Jargon terms are observed in 5.2\% of the utterances; when short utterances containing less than 5 words are removed from consideration, the percentage is fairly significant (11.6\%).

Our collection of domain terminology will be a valuable resource to investigate a variety of research questions regarding domain adaptation.
Importantly, if a summarizer performs better when jargon terms are excluded, it indicates domain terminology may have only limited impact on determining utterance salience, or the summarizer has been ineffective in using domain knowledge.
Conversely, if the summarizer performs less well, domain terminology is considered essential and it is important for speech recognizers to correctly transcribe these terms to avoid any loss in summarization performance.
In what follows, we describe our meeting summarizer and examine how domain terms are processed by a modern tokenizer.

\begin{wraptable}{r}{0.53\textwidth}
\setlength{\tabcolsep}{5pt}
\renewcommand{\arraystretch}{1.1}
\centering
\begin{fontppl}
\begin{footnotesize}
\begin{tabular}{|l|l|}
\hline
\textbf{Jargon Term} & \textbf{Tokenization} \\
\hdashline
SmartKom system & \textcolor{magenta}{\textbf{smart-ko-m}} system\\
discourse annotations & discourse \textcolor{magenta}{\textbf{ann-ota-tions}}\\
situational context factors & situation-al context factors\\
modifiers, auxiliaries & mod-ifiers , aux-ilia-ries\\
JavaBayes belief-net & \textcolor{magenta}{\textbf{java-bay-es}} belief - net\\
a real wizard system & a real wizard system\\
the L\_D\_C & the \textcolor{magenta}{\textbf{l \_ d \_ c}}\\
the near field mikes & the near field \textcolor{magenta}{\textbf{mike-s}}\\
\hline
\hline
\multicolumn{2}{|l|}{\textbf{Utterance with Jargon}}\\
\hdashline
\multicolumn{2}{|l|}{she wanted to display the \textcolor{magenta}{\textbf{stylized F\_ zeroes,}} I think}\\
\multicolumn{2}{|l|}{they're called?}\\
\hline
\multicolumn{2}{|l|}{\textbf{Utterance without Jargon}}\\
\hdashline
\multicolumn{2}{|l|}{she wanted to display the \textcolor{magenta}{\textbf{[MASK]}} I think they're called?}\\
\hline
\end{tabular}
\end{footnotesize}
\end{fontppl}
\caption{
An example showing how jargon terms are processed by a modern tokenizer, WordPiece. 
E.g., \emph{smart-ko-m} means the jargon \emph{SmartKom} was split into three tokens.
Moreover, our method allows jargon to be masked-out of the utterances for summarization.
}
\label{tab:tokenize} 
\end{wraptable}

\section{Meeting Summarization}
\label{sec:approach}

The very first step that one must take to build a meeting summarizer is tokenization, which transforms an input utterance to a sequence of \emph{sub-word units}.
WordPiece~\cite{6289079} and BPE~\cite{sennrich-etal-2016-neural} are two modern methods for tokenization.
We use WordPiece that has a total vocabulary of 30,522 sub-words.
The method builds a vocabulary of the desired size by iteratively combining word parts into a sub-word if doing so increases the language model likelihoods.
Given the vocabulary and any input word, it uses a greedy longest-match-first algorithm to tokenize it into sub-word units; the longest sub-word will be matched first.
We provide example tokenization outputs in Table~\ref{tab:tokenize}.

\begin{table*}
\setlength{\tabcolsep}{6pt}
\renewcommand{\arraystretch}{1.1}
\centering
\begin{fontppl}
\begin{footnotesize}
\begin{tabular}{|ll|ccc|ccc|ccc|}
\hline
& & \multicolumn{3}{c|}{\textbf{ROUGE-1}} & \multicolumn{3}{c|}{\textbf{ROUGE-2}} & \multicolumn{3}{c|}{\textbf{ROUGE-SU4}}\\
& & P(\%) & R(\%) & F(\%) & P(\%) & R(\%) & F(\%) & P(\%) & R(\%) & F(\%)\\
\hline
\hline
\multirow{3}{*}{\textsc{Human}} & \cite{xie-liu-2010-using} & -- & -- & {\cellcolor[gray]{.9}}\textbf{69.1} & -- & -- & 33.3 & -- & -- & --\\
& Ours (w/o Jargon) & {59.5} & 63.8 & 59.9 & 32.9 & 34.4 & 32.8 & 33.5 & 35.3 & 33.6\\
& Ours (w/\;\; Jargon) & 57.0 & {70.6} & 60.7 & {34.9} & {43.0} & {\cellcolor[gray]{.9}}\textbf{37.1} & {34.9} & {43.2} & {\cellcolor[gray]{.9}}\textbf{37.2}\\ 
\hdashline
\multirow{3}{*}{\textsc{ASR}} & \cite{shang-etal-2018-unsupervised} & 27.6 & 36.3 & 31.0 & 4.4 & 5.6 & 4.8 & 9.9 & 13.5 & 11.3\\
& Ours (w/o Jargon) & 41.7 & 55.1 & {\cellcolor[gray]{.9}}\textbf{46.8} & 15.1 & 20.5 & 17.2 & 18.7 & 25.3 & 21.3\\
& Ours (w/\;\; Jargon) & 39.7 & 57.5 & 46.6 & 15.1 & 21.9 & {\cellcolor[gray]{.9}}\textbf{17.7} & 18.2 & 26.5 & {\cellcolor[gray]{.9}}\textbf{21.4}\\
\hline
\end{tabular}
\end{footnotesize}
\end{fontppl}
\caption{
Results of our summarizer on the ICSI test set. 
We report ROUGE scores for our summarizer, with and without using jargon, and contrast it with strong baseline systems~\cite{xie-liu-2010-using,shang-etal-2018-unsupervised}. 
Experimental results on human transcripts and speech recognition outputs (ASR) suggest that our model performs on par with prior state of the art. 
}
\label{tab:results_summ} 
\end{table*}

We show that most domain terminology can be properly processed by the WordPiece tokenizer.
There are two immediate issues that require attention. 
First, it has considerable difficulties processing infrequent entities and terms, e.g., \emph{smart-ko-m}, \emph{java-bay-es} and \emph{ann-ota-tions} are not well tokenized.
Moreover, entities such as ``LDC'' need to be spelled out, the tokenizer transforms it into three individual letters, thus losing the original meaning.

Our meeting summarizer takes as input an utterance and outputs a binary label indicating if the utterance should be included in the summary.
Due to data scarcity, we refrain from using sequential prediction or a more sophisticated approach that may overfit, but focus primarily on demonstrating the impact of domain terminology on model performance.
Our summarizer is based on BERT-\textsc{large} that contains 24 layers of Transformer blocks, 16 attention heads and 1024-dimensional hidden vectors~\cite{devlin-etal-2019-bert}.
The top-layer hidden vector of the \textsc{[cls]} token is used as the representation of the input utterance.
We apply a linear and a softmax layer to predict a binary label.
Importantly, jargon terms can be masked-out of the input utterances by replacing each term with \textsc{[mask]} token prior to training (Table~\ref{tab:tokenize}).
The method thus employs a single architecture to assess model performance with and without jargon.

We train the summarizer on 38,657 utterances from 54 meetings; each meeting was annotated by a single annotator.
Utterances containing less than 5 words are removed from consideration.
The summarizer is evaluated on the standard test set containing 6 meetings; 
each of these meetings have been annotated by three annotators~\cite{Carenini:2011}.
Our experiments are performed on human transcripts and ASR outputs, respectively, the latter are acquired from the SRI speech recognizer.
In the following, we discuss our findings in terms of how domain terminology affects summarization.

\section{Results and Analysis}
\label{sec:results}

Our experimental results are presented in Table~\ref{tab:results_summ}.
We evaluate against two strong baseline systems.
Xie and Liu~\shortcite{xie-liu-2010-using} describe an extractive meeting summarizer utilizing maximum marginal relevance and speech-specific features.
Shang et al.~\shortcite{shang-etal-2018-unsupervised} introduce a graph framework to group utterances into clusters, perform multi-sentence compression then selection under a budget constraint.
Our experiments show that, despite its simplicity, our meeting summarizer can outperform or perform on par with prior state of the art, showing a remarkable advancement of pretrained deep models in the meeting domain.

We observe that summarizing with jargon terms yields substantially better performance (an absolute gain of +4.3\% R-2 F-score) on human transcripts, comparing to the alternative that masks jargon out of input utterances.
The performance gap has narrowed on ASR transcripts, as domain terminology contains infrequent entities and terms, which are subject to transcription errors.
Our findings suggest that domain terminology plays a significant role in determining utterance salience.
Its impact on summarization and other downstream meeting applications should not be underestimated.

In Table~\ref{tab:summ_length}, we assess the model performance on human transcripts, using jargon or not during training, and generate output summaries of varying length. 
We rank the utterances by their confidence scores and select a portion of them. \emph{Gold} uses the length of ground-truth summaries.
We show the precision, recall and F-scores of our classifier, ROUGE~\cite{lin-2004-rouge} and BERTScore~\cite{Zhang2020BERTScore} for summaries.\footnote{The hash code for BERTScore is \textsf{xlnet-base-cased\_L12\_no-idf\_version=0.3.4(hug\_trans=2.5.1)-rescaled}}
We find that across all lengths and evaluation metrics, summarizing with jargon can lead to a performance boost for meeting summarization.
While this work has primarily experimented with the ICSI corpus, the results are sufficiently substantial that we expect them to hold over similar meeting corpora.

\begin{table*}
\setlength{\tabcolsep}{5pt}
\renewcommand{\arraystretch}{1.1}
\centering
\begin{fontppl}
\begin{footnotesize}
\begin{tabular}{|l|ccc|cc|c||ccc|cc|c|}
\hline
& \multicolumn{6}{c||}{\textbf{Without Jargon}} & \multicolumn{6}{c|}{\textbf{With Jargon}}\\
Summ & \multicolumn{3}{c|}{Classifier} & \multicolumn{3}{c||}{Summarizer} & \multicolumn{3}{c|}{Classifier} & \multicolumn{3}{c|}{Summarizer}\\
Length & P(\%) & R(\%) & F(\%) & R-1 & R-2 & BERTScore & P(\%) & R(\%) & F(\%) & R-1 & R-2 & BERTScore\\
\hline
\hline
5\% & 37.4 & 10.0 & 15.8 & 36.4 & 18.5 & 54.0 & 39.3 & 10.5 & 16.6 & 37.2 & 19.1 & 53.6\\
10\% & 34.3 & 18.3 & 23.8 & 51.9 & 26.2 & 57.6 & 37.1 & 19.8 & 25.8 & 53.8 & 29.7 & 57.3\\
15\% & 32.2 & 25.7 & 28.6 & 58.0 & 29.7 & 59.8 & 37.8 & 30.2 & 33.5 & 60.9 & 35.5 & 60.3\\
20\% & 32.0 & 34.1 & 33.0 & 60.3 & 33.5 & {\cellcolor[gray]{.9}}\textbf{62.0} & 35.1 & 37.4 & 36.2 & 62.0 & 37.1 & {\cellcolor[gray]{.9}}\textbf{62.1}\\
\hdashline
Gold & 33.5 & 33.5 & {\cellcolor[gray]{.9}}\textbf{33.5} & {\cellcolor[gray]{.9}}\textbf{61.7} & {\cellcolor[gray]{.9}}\textbf{33.5} & 61.4 & 37.0 & 37.0 & {\cellcolor[gray]{.9}}\textbf{37.0} & {\cellcolor[gray]{.9}}\textbf{63.7} & {\cellcolor[gray]{.9}}\textbf{38.6} & 61.3\\
\hline
\end{tabular}
\end{footnotesize}
\end{fontppl}
\caption{Results of our meeting summarizer, using jargon or not, while varying the length of output summaries. 
}
\label{tab:summ_length} 
\end{table*}

\section{Related Work}
\label{sec:related}

Generating meeting summaries is a challenging problem with a great application potential.
A significant number of techniques have been attempted in the past, including extraction of utterances and keyphrases from transcripts~\cite{galley-2006-skip,murray-carenini-2008-summarizing,liu-etal-2009-unsupervised,gillick2009} and taking advantage of prosodic and speaker-related features~\cite{Maskey05comparinglexical,zhu-etal-2009-summarizing,DBLP:conf/slt/ChenM12}.
As spoken utterances are verbose with low information density, some methods further compress and merge utterances~\cite{Liu:2013:IEEETrans,wang-cardie-2013-domain,mehdad-etal-2013-abstractive}.
Despite these valuable contributions, a closer investigation remains necessary to develop an understanding of how domain terminology affects meeting summarization performance.

Recent years have seen a renewed interest in summarizing meeting transcripts~\cite{shang-etal-2018-unsupervised,zhu2020endtoend,tardy-etal-2020-align} and other types of online and transcribed conversations~\cite{goo2018abstractive,yuan2019abstractive,gliwa-etal-2019-samsum}.
In particular, Tardy et al.~\shortcite{tardy-etal-2020-align} create a corpus containing 22 public meetings including their automatic transcriptions from audio recordings and meeting reports written by a professional.
Li et al.~\shortcite{li-etal-2019-keep} develop a multi-modal hierarchical attention mechanism for abstractive summarization, where attention is applied to topics, utterances and words to narrow the focus to salient content; their experiments were performed the AMI corpus, thus results are not directly comparable.
Our work excludes prosodic and speaker-related features to focus solely on domain terminology.
It provides a new baseline for future research toward building effective meeting summarizers.

\section{Conclusion}
\label{sec:conclusion}

We seek to better understand how domain terminology impacts meeting summarization performance in the context of neural extractive summarization. 
We solicit quality annotations from expert annotators to compile a list of jargon terms from a sizable meeting corpus, which is a valuable resource to investigate a variety of research questions regarding domain adaptation.
Our extensive experiments show that domain terminology has a substantial impact on summarization performance that should not be neglected.
Future work may address the questions of how to obtain domain terminology in a semi-automatic way and inject domain knowledge into a meeting summarization system.

\section*{Acknowledgements}
We are grateful to the anonymous reviewers for their insightful feedback.
This research was supported in part by the NSF grant IIS-1909603.
Xiaojin Dai was supported by NSF DUE-1643835.

\bibliographystyle{coling}
\bibliography{anthology,fei,more}

\end{document}